\documentclass[letterpaper, 10 pt, conference]{ieeeconf}

\usepackage[T1]{fontenc}
\usepackage{cite}
\usepackage{amssymb,amsfonts}
\usepackage{blindtext}
\makeatletter
\let\NAT@parse\undefined
\makeatother
\usepackage{hyperref}
\usepackage{graphicx}
\usepackage{textcomp}
\usepackage{mathtools}
\usepackage{changes}
\usepackage[font=footnotesize,labelformat=simple]{subcaption}
\usepackage{algorithm}
\usepackage{xcolor}
\usepackage{color, soul}
\usepackage{amsmath}
\usepackage{booktabs}
\usepackage{multirow}
\usepackage{xstring}
\usepackage{hhline}
\usepackage{kotex}
\usepackage{siunitx}
\usepackage{comment}
\usepackage{physics}
\usepackage{tikz}
\usepackage{cleveref}
\usepackage{gensymb}
\usepackage{threeparttable}
\usepackage{caption}
\usepackage{url_bib_utils_short}
\usepackage{bm}
\usepackage{algpseudocode}
\usepackage{algorithm}
\usepackage{float}
\usepackage{placeins}
\usepackage{listings}
\usepackage{threeparttable}

\definecolor{codegreen}{rgb}{0,0.6,0}
\definecolor{codegray}{rgb}{0.5,0.5,0.5}
\definecolor{codepink}{RGB}{252, 142, 172}
\definecolor{codepurple}{rgb}{0.58,0,0.82}
\definecolor{backcolour}{RGB}{245,245,245}
\lstdefinestyle{mystyle}{
    backgroundcolor=\color{backcolour},   
    commentstyle=\color{magenta},
    keywordstyle=\color{blue},
    numberstyle=\tiny\color{codegray},
    stringstyle=\color{codepurple},
    basicstyle=\fontfamily{\ttdefault}\footnotesize,
    breakatwhitespace=false,         
    breaklines=true,                 
    keepspaces=true,    
    frame=single,
    numbersep=5pt,                  
    showspaces=false,                
    showstringspaces=false,
    showtabs=false,                  
    tabsize=2,
    classoffset=1, %
    keywordstyle=\color{violet},
    classoffset=0,
}
\crefformat{section}{\S#2#1#3} 
\crefformat{subsection}{\S#2#1#3}
\crefformat{subsubsection}{\S#2#1#3}

\def\eqref#1{Eq.~(\ref{#1})}

\definecolor{pr}{RGB}{0, 0, 0}
\definecolor{greenfoot}{RGB}{126,170,85}
\definecolor{orangefoot}{RGB}{222,131,68}

\IEEEoverridecommandlockouts                              

\newcommand*{\acro}{LocoVLM\xspace}
\newcommand*{\motdesc}{motion descriptor\xspace}
\newcommand*{\motdescs}{motion descriptors\xspace}
\newcommand*{\llm}{GPT-4o~\cite{achiam2023gpt}\xspace}

\newcommand*{\vl}{vision-language\xspace}
\newcommand*{\gaitencoding}{gait phase encoding\xspace}
\newcommand*{\retrieval}{mixed-precision retrieval\xspace}
\def\lang#1{\texttt{#1}}

\DeclareMathOperator*{\argmax}{arg\,max}

\DeclareMathOperator*{\softmax}{softmax}
\DeclareMathOperator*{\cossim}{cossim}

\IEEEoverridecommandlockouts  

\title{\textbf{\acro: Grounding Vision and Language for Adapting Versatile Legged Locomotion Policies}}

\author{
I Made Aswin Nahrendra, Seunghyun Lee, Dongkyu Lee, and Hyun Myung$^{*}$
\thanks{
The authors are with the School of Electrical Engineering, Korea Advanced Institute of Science and Technology (KAIST), Daejeon 34141, Republic of Korea (e-mail: \texttt{\{anahrendra, kevin9709, dklee98, hmyung\}@kaist.ac.kr}).
}
\thanks{
$^{*}$Corresponding author: Hyun Myung.
}
}

\begin{document}

\maketitle

\begin{abstract}
    Recent advances in legged locomotion learning are still dominated by the utilization of geometric representations of the environment, limiting the robot's capability to respond to higher-level semantics such as human instructions. To address this limitation, we propose a novel approach that integrates high-level commonsense reasoning from foundation models into the process of legged locomotion adaptation. Specifically, our method utilizes a pre-trained large language model to synthesize an instruction-grounded skill database tailored for legged robots. A pre-trained vision-language model is employed to extract high-level environmental semantics and ground them within the skill database, enabling real-time skill advisories for the robot. To facilitate versatile skill control, we train a style-conditioned policy capable of generating diverse and robust locomotion skills with high fidelity to specified styles. To the best of our knowledge, this is the first work to demonstrate real-time adaptation of legged locomotion using high-level reasoning from environmental semantics and instructions with instruction-following accuracy of up to $87\%$ without the need for online query to on-the-cloud foundation models.
\end{abstract}

\IEEEpeerreviewmaketitle

\section{Introduction}
Legged robots hold immense potential for real-world applications owing to their ability to navigate complex terrains and environments. Over the past decade, they have become increasingly developed for their potential to assist humans in various tasks such as inspection, last-mile delivery, and search-and-rescue missions~\cite{bouman2020autonomous,lee2024learning,jacoff2023taking}. Recent advancements in deep reinforcement learning~(RL) have significantly accelerated research in this domain. RL, in particular, has shown promising results in generating robust locomotion policies capable of adapting to diverse terrains and tasks~\cite{kumar2021rma,lee2020learning,rudin2022learning,ji2022concurrent,nahrendra2023dreamwaq,rudin2022advanced}. A critical challenge in legged locomotion lies in perceiving the environment and acting accordingly while fully utilizing the robot's physical capabilities.

Typically, legged robot locomotion is governed by controllers that rely on geometric representations of the environment. These representations are derived either explicitly from exteroceptive sensors or implicitly from proprioceptive sensors~\cite{miki2022learning,hoeller2023anymal,cheng2024extreme,nahrendra2024obstacle}. However, geometric representations fail to capture environmental semantics, such as object affordances, social awareness, or task-specific details. Understanding environmental semantics is essential for robots to synthesize interactive and skillful behaviors, which are crucial for executing more complex high-level tasks~\cite{tang2023saytap,rana2023sayplan,ahn2022can,ginting2024saycomply}.
\begin{figure}[!t]
	\centering 
	\includegraphics[width=0.9\linewidth]{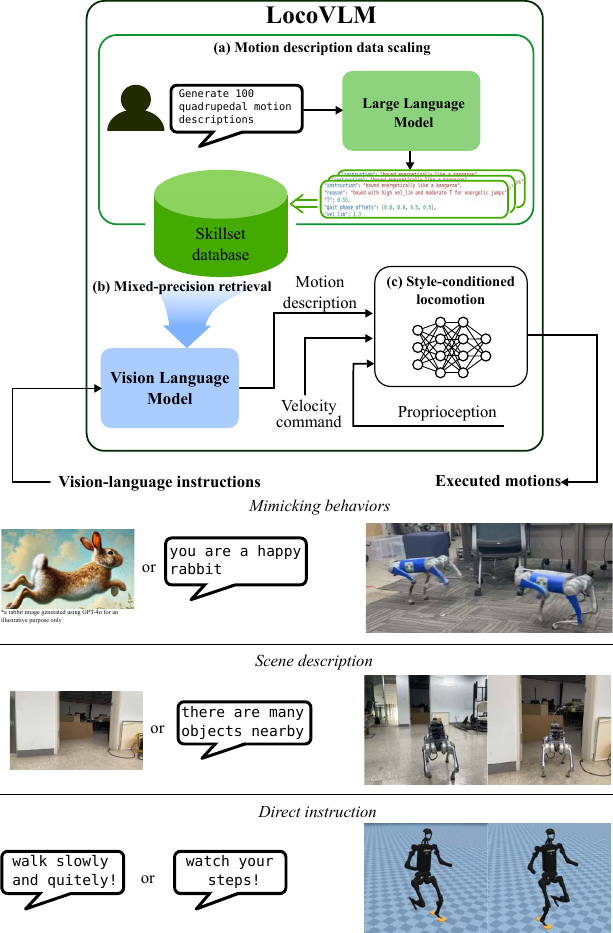}
	\caption{LocoVLM receives vision-language instructions as its input and hierarchically grounds them into versatile locomotion skills. 
    (a)~An LLM is used to scale up motion descriptor data generation and store it into a skill database. 
    (b)~During inference, a VLM retrieves the most relevant motion descriptor from the database using the proposed mixed-precision retrieval mechanism to give a style reference to the locomotion controller. 
    (c)~Finally, a pre-trained style-conditioned locomotion controller executes the robot’s motion to realize the given vision-language instructions.}
	\label{figure:teaser}
\end{figure}
Recent advancements in foundation models, such as large language models~(LLMs) and vision-language models~(VLMs), have created opportunities to integrate higher-level reasoning with low-level robot control by using \vl observations~\cite{brohan2022rt,brohan2023rt,belkhale2024rt}. However, the advantages of modern foundation models are often hindered by challenges in integrating these large models with control policies that require real-time decision-making~\cite{mirchandani2023large,kim2024survey}. 


We propose \acro, a novel framework for legged locomotion that brings the vast knowledge of foundation models into the field of legged locomotion. \acro grounds image and language inputs to adapt versatile locomotion policies in real time, yielding a robust and interactive legged locomotion system as shown in the accompanying video.\footnote{\url{https://locovlm.github.io}}



Fig.~\ref{figure:teaser} provides an overview of our proposed \acro framework, which can be viewed as a hierarchical system following a teacher-student paradigm for the high-level policy. The LLM functions as a high-level teacher policy, generating high-level commands for the robot, while the VLM serves as a high-level student policy that extracts a subset of domain knowledge from the teacher policy. The low-level policy is the style-conditioned locomotion controller that executes the given motions based on the high-level commands.

In summary, the contributions of this paper aim to address the following key challenges in integrating \vl models with legged robot locomotion:
\begin{enumerate}
    \item \textbf{Robust style-conditioned locomotion.} A locomotion policy learning framework that efficiently conditions locomotion styles while maintaining robustness, enabling the robot to accurately follow semantic instructions without compromising stability.
    \item \textbf{Foundation model knowledge distillation.} A scalable data generation pipeline that utilizes an LLM to generate a database of instructions and their corresponding executable \motdesc, allowing real-time inference and semantic adaptation of the robot's locomotion policy without requiring in-the-loop queries to an online LLM.
    \item \textbf{Fast and accurate retrieval.} A \vl grounding approach that facilitates real-time retrieval of \motdesc from the database using a pre-trained VLM, enabling the robot to adapt its locomotion policy based on human instructions or robot-centric observations.
    \item \textbf{Generalization.} An analysis of the framework's generalization capabilities across different tasks and embodiments, highlighting its potential for further applications in real-world scenarios.
\end{enumerate}

\section{Related Work}\label{section:related_work}
\subsection{Locomotion Skill Control}
Recent advancements in learning methods for legged locomotion have enabled robust and skillful control, significantly improving the robot's adaptability and versatility. The learning of these locomotion skills is generally guided by reference commands in the form of base velocity~\cite{lee2020learning,kumar2021rma,miki2022learning, nahrendra2023dreamwaq}, foot trajectory~\cite{tsounis2020deepgait}, or body pose~\cite{rudin2022advanced,hoeller2023anymal}. Recent works in learning-based locomotion control have also nurtured the learning of controllable skills using heuristics defined in the reward functions. These heuristic rewards have facilitated the learning of gait style-conditioned~\cite{margolis2023walk,kim2024learning,bellegarda2024allgaits,humphreys2024learning} or contact-scheduled~\cite{tang2023saytap,zhang2024wococo} policies. However, these heuristic reward functions often undermines the controller's robustness, as the reward design may tend to satisfy the heuristics rather than maintaining robustness against disturbances.

\subsection{LLM as Robot Policies}
The success of LLMs in reasoning and generating human-like text has sparked interest in using them as robot policies. A common approach runs the LLM in the control loop to generate executable actions, such as code~\cite{liang2023code}, sub-task plans~\cite{rana2023sayplan,mandi2024roco}, rewards~\cite{yu2023language}, or high-level commands~\cite{tang2023saytap,jiao2023swarm,mirchandani2023large}. However, deploying LLMs for real-time control remains challenging, especially in field deployments, where a dedicated cloud connectivity is often unavailable. As a result, most deployments are constrained to controlled environments with dedicated networks~\cite{tang2023saytap,ginting2024saycomply,rana2023sayplan}, or must trade off smooth real-time performance~\cite{mirchandani2023large,cheng2024navila}.
\subsection{LLM as a Data Generator}
Another paradigm for utilizing LLMs is to employ them as data generators. The motivation behind this approach is: \textit{if LLMs can generate human-like texts, can they also annotate or code like humans?} This idea has been recently explored to scale up data generation for training robot policies with minimal human effort in data collection~\cite{ha2023scaling,yu2024learning} and to accelerate code composition for training environment generation~\cite{ma2024eureka,ma2024dreureka,liang2024environment}. Leveraging LLMs as data generators can rapidly scale up data for various tasks, reducing the manual effort required for data annotation.

\section{Methodology}
\subsection{Versatile Quadrupedal Locomotion}
\subsubsection{Style-Conditioned Locomotion Policy}
We trained a style-conditioned locomotion policy using a blind velocity-conditioned locomotion learning framework~\cite{kumar2021rma,lee2020learning,ji2022concurrent,margolis2023walk,nahrendra2023dreamwaq}, augmented with a style parameter vector that parameterizes gait style using a gait cycle duration $T$ and gait phase offsets $\psi\in\mathbb{R}^4$. A \gaitencoding vector was used as a clock input for the policy, defined as $\bm{\phi}(t) = [\sin\left(2\pi \frac{t}{T}\right),  \cos\left(2\pi \frac{t}{T}\right)]$, where $t$ denote the current time step.
\subsubsection{Compliant Contact Tracking}

\begin{figure}{}
	\centering 
    \includegraphics[width=0.8\linewidth]{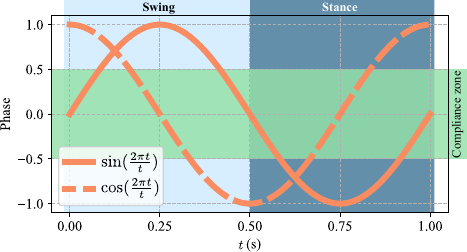}
	\caption{Gait phase encoding for a cycle duration of $T\!=\!1$. The \gaitencoding vector is a two-dimensional representation of the current phase in the gait cycle.}
	\label{figure:gait_encoding}
\end{figure}

Style-conditioned contact tracking often compromises locomotion robustness to enhance style accuracy. Therefore, we propose a compliant contact tracking method that allows accurate gait tracking while preserving the robustness of the locomotion controller. This behavior is achieved by incorporating a compliance term into the contact tracking reward function, which imposes zero penalty when the foot is not in the correct contact state within the compliance threshold. The compliance term is defined as:
\begin{equation}
    \phi^\text{comply}_\text{error} = 
    \begin{cases}
        0, & \text{if}~\phi_\text{error} \leq \delta, \\
        \phi_\text{error}, & \text{otherwise},
    \end{cases}
    \label{equation:compliance}
\end{equation}
where $\delta$ is the compliance threshold, and $\phi_\text{error}$ represents the error between the desired and actual contact states. The desired contact states for the swing and stance phases are zero and one, respectively. Fig.~\ref{figure:gait_encoding} illustrates the \gaitencoding for $T=1$ with $\delta=0.5$. Within the compliance zone (green shade), the policy is not rewarded for tracking the cycle phase (orange curves), allowing it to compliantly adapt to disturbances.

\subsection{Scaling Up Motion Description Data}
We utilize an LLM to generate a database of instructions and \motdescs for the robot's locomotion policy. By creating a database of instructions and their corresponding \motdescs, we distill the commonsense knowledge of the LLM into a structured instruction set tailored for the robot's locomotion policy.

\subsubsection{Instruction Description Generation}
\begin{figure}[!ht]
	\centering 
	\includegraphics[width=\linewidth]{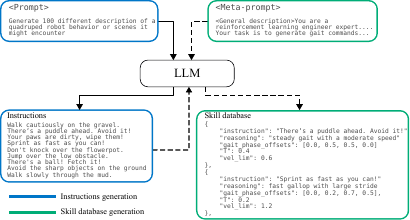}
	\caption{Offline skill database generation pipeline. The LLM firstly generates instructions, which are categorized into mimicking behaviors, scene responses, and direct instructions. These instructions are then passed to a meta-prompt to generate contents for the skill database.}
	\label{figure:datagen}
	\vspace{-5pt}
\end{figure}
We propose a two-stage data generation pipeline to efficiently scale up the data generation process as shown in Fig.~\ref{figure:datagen}. This pipeline offers two primary benefits: (1) it mitigates the maximum token length constraint of the LLM, and (2) it enables the LLM to generate diverse and structured motion descriptors by passing the generated instructions to a meta-prompt that produces the motion descriptors. We used the \llm model as the LLM.

In the first stage, we prompted the LLM to generate a set of brief instructions and/or scene descriptions $\mathbf{I}$. These instructions are categorized into three types: (1)~mimicking behaviors, (2)~responding to a scene, and (3)~following direct instructions. 

In the second stage, skill database was generated by prompting the LLM to generate $n$ instructions based on the instruction category. We observed that the LLM produces more diverse yet structured instructions when prompted categorically. This categorical prompting approach prevents the LLM from hallucinating false instructions and ensures that the generated instructions are relevant to the category. To comply with the maximum token length constraint of the LLM, we set $n\!=\!100$ for each generation process. The system prompts used for instruction description generation are supplemented in Appendix~\ref{appendix:instruction_prompt}.

\subsubsection{Instruction-Grounded Motion Description}

\begin{figure}[!t]
	\centering 
	\includegraphics[width=\linewidth]{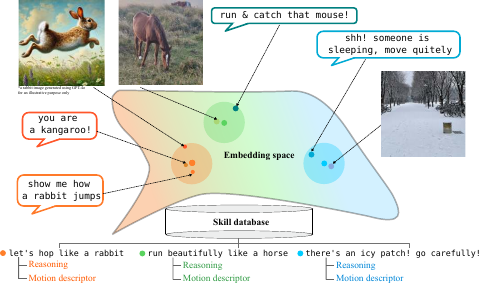}
	\caption{Skill database retrieval process. The instruction query is encoded by a VLM into a learned embedding space. A VLM is used instead of a sentence encoder to enable multimodal retrieval from text or image inputs. The closest instruction embedding in the database is retrieved to obtain the corresponding \motdesc.}
	\label{figure:skill_database}
\end{figure}

The LLM-generated instructions and \motdescs are stored in a skill database $\mathcal{D}$. During deployment, a pre-trained VLM encodes the input instruction, provided as text or an image, into an embedding space. The closest instruction embedding in the database is then retrieved to obtain the corresponding \motdesc, as illustrated in Fig.~\ref{figure:skill_database}.

Each entry in the database is a tuple $d=(\mathcal{I}, \mathbf{M})$, where $\mathcal{I} \in \mathbf{I}$ is the instruction text and $\mathbf{M}$ is the motion descriptor. The motion descriptor is a tuple defined as follows:
\begin{equation}
    \mathbf{M} = ( T, \; \psi, \; v_x^\text{limit} )
\end{equation}
where $\psi$ is the gait phase offsets and $v^\text{limit}_x$ is the maximum velocity along the $x$-axis. Without loss of generality, we use only the velocity limit along the $x$-axis in the motion descriptor. However, extending the descriptor to include velocity limits along the $y$-axis and the yaw rate is straightforward and could benefit the generation of more diverse motions. In this paper, we limit the \motdesc to three elements, representing the minimum parameters required to synthesize various locomotion styles.

\subsubsection{Prompted Reasoning for Motion Description Generation}\label{section:method_reasoning}
Generating low-level motion descriptions from direct instructions (e.g., ``\lang{walk forward}'') is relatively straightforward, as these can be directly mapped to parameters such as gait type, gait cycle period, and velocity limit. However, vague instructions (e.g., ``\lang{move like a hippo}'') or contextual cues (e.g., ``\lang{you are in a library}'') pose greater challenges.

To circumvent this issue, we introduce a prompted reasoning method that guides the LLM in generating diverse, executable skills. The key idea is to translate high-level instructions into detailed and technical descriptions. This is achieved by prepending a system prompt that asks the LLM to first produce reasoning before outputting a motion description. As shown in Fig.~\ref{figure:skill_database}, each skill entry consists of an instruction, its reasoning, and the resulting \motdesc. Details of the system prompt for the skill generation with prompted reasoning is provided in Appendix~\ref{appendix:skill_prompt}.

\subsection{Vision-Language Model as a Motion Advisor}
\subsubsection{Mixed-Precision Retrieval}\label{section:mixed_retrieval}

During deployment, \acro retrieves the most similar instruction from the skill database and forwards it to the locomotion policy. We used the encoder of a pre-trained BLIP-2 model~\cite{li2023blip} to extract embeddings from both text and image queries. These embeddings are then compared with the instructions in the database to identify the closest match.

A key challenge in database scaling is retrieval efficiency. Accurate retrieval using BLIP-2’s image-text-matching (ITM) head requires exhaustively comparing the query against all database entries. In contrast, computing cosine similarity between query and instruction embeddings is more scalable but less accurate, as it does not leverage the ITM head’s pre-trained feature fusion~\cite{li2023blip}.

Therefore, we introduce a \retrieval method that combines the best of both schemes. Our \retrieval method breaks down the retrieval process into two stages. The problem of retrieving an instruction given a query is formulated as follows:
\begin{equation}
	\mathcal{I}^* = \argmax_{\mathcal{I} \in \mathcal{D}} \text{sim} \big(\mathcal{I}_\text{query}, \mathbf{I}\big),
	\label{eq:retrieval}
\end{equation}
where $\mathcal{I}_\text{query}$ represents the instruction query, $\mathcal{I}^*$ is the retrieved instruction, $\mathcal{D}$ is the skill database, and $\text{sim}()$ is the similarity metric. A pseudocode summarizing the proposed mixed-precision retrieval method is provided in Algorithm~\ref{algo:retrieval}.

\begin{figure}
	\begin{minipage}{0.5\textwidth}
	\begin{algorithm}[H]
		\caption{Mixed-Precision Retrieval}
		\label{algo:retrieval}
		\footnotesize
		\begin{algorithmic}[1]
			\State \textbf{Input:} $\mathcal{I}_\text{query}$, $\mathcal{D}$, $f_\text{BLIP}$, $f_\text{ITM}$, $K$
			\State $\mathbf{I}^K \gets \argmax_{\mathcal{I} \in \mathcal{D}} \cossim \big(f_\text{BLIP}(\mathcal{I}_\text{query}), f_\text{BLIP}(\mathbf{I})\big)$
			\State $p_\text{1}(\mathbf{I}^K) \gets \softmax \left(\cossim \big(f_\text{BLIP}(\mathcal{I}_\text{query}), f_\text{BLIP}(\mathbf{I}^K)\big)\right)$
			\For{$\mathcal{I}^k \in \mathbf{I}^K$}
				\State $p_\text{2}(\mathcal{I}_k) \gets \softmax \big( f_\text{ITM}(\mathcal{I}_\text{query}, \mathcal{I}_k) \big)$
			\EndFor
			\State $\mathcal{I}^* \gets \argmax_{\mathcal{I}_k} \left( p_\text{1}(\mathbf{I}^K) + p_\text{2}(\mathbf{I}^k) \right)$
			\State \textbf{Output:} $\mathcal{I}^*$
		\end{algorithmic}
	\end{algorithm}
	\end{minipage}
\end{figure}
\subsubsection{Text as Image Helps Sentence Understanding}
State-of-the-art VLMs are generally trained on large image-text datasets using contrastive methods~\cite{radford2021learning,li2022blip,li2023blip}, which align image-text pairs but ignore relationships between text pairs. To address this limitation, we hypothesize that representing text as an image can improve VLM performance. Specifically, we render the text query as a \textit{text image}, a white background with foreground text—using a simple plotting tool like Matplotlib~\cite{Hunter:2007}. This image is then fed into the VLM to perform image-based text retrieval by comparing it with instruction images in the skill database. This approach leverages the VLM’s strength in image-text matching to maximize \acro’s retrieval performance.


\section{Experiments}~\label{section:experiments}
We evaluate LocoVLM through a series of experiments to validate our contributions by addressing the following questions:
\begin{enumerate}
    \item \textbf{Robust style-conditioned locomotion.} Is the proposed compliant contact tracking method capable of executing various locomotion styles while maintaining robustness?~(Section~\ref{section:results_gait_tracking})
    \item \textbf{Foundation model knowledge distillation.} Is the LLM-generated database of instructions and their corresponding executable \motdescs effective for real-time inference and adaptation of the robot's locomotion policy without requiring the LLM during deployment?~(Section~\ref{section:results_reasoning})
    \item \textbf{Fast and accurate retrieval.} Can the proposed retrieval method quickly and accurately retrieve the most relevant \motdesc from the database using a pre-trained VLM?~(Section~\ref{section:results_retrieval})
    \item \textbf{Generalization.} How well does the proposed \acro generalize across different tasks and embodiments without further fine-tuning?~(Section~\ref{section:results_humanoid})
\end{enumerate}

\subsection{System Setup}\label{experiments:system_setup}
\subsubsection{Locomotion Controller}
The locomotion controller for a Unitree Go1 was trained using the proximal policy optimization~(PPO) algorithm~\cite{schulman2017proximal} with an asymmetric actor-critic implementation and state estimation, following the methodologies of~\cite{ji2022concurrent} and~\cite{nahrendra2023dreamwaq}. The policy was trained in a custom implementation of the training environment built using the Isaac Gym simulator~\cite{makoviychuk2021isaac}. Several domain parameters were randomized to ensure sim-to-real robustness, including robot mass, center of mass, motor stiffness, damping, terrain friction, and system delays.

The locomotion policy was deployed on the Unitree Go1's onboard Jetson Xavier NX board. The policy was operated at $50~\text{Hz}$, sending joint angle commands to the robot's motor controllers that convert the commands to motor torques at $200~\text{Hz}$.

\subsubsection{VLM Inference}
The VLM module was run on a separate laptop equipped with an NVIDIA GeForce RTX 3070~Ti GPU. All input instructions, whether in text or image format, were transmitted to the VLM module via a robot operating system~(ROS) network. The retrieved motion descriptor was subsequently sent to the robot through the same ROS network. We observed no performance degradation due to network latency, as the VLM advisor and locomotion controller operate asynchronously. Furthermore, the data transmission and VLM inference time were negligible, amounting to less than $100~\text{ms}$.


\subsection{Gait Tracking Performance}\label{section:results_gait_tracking}
\begin{table}[t!]
    \centering
    \begin{threeparttable}
    \footnotesize
    \caption{Gait tracking performance of the compliant contact tracking method with different compliance threshold value of $\delta$.}
    \label{table:gait_tracking}

    \begin{tabular}{llccc}
        \hline
        \multicolumn{1}{c}{\multirow{2}{*}{Gait}} 
        &\multicolumn{1}{c}{\multirow{2}{*}{$\delta$}} 
        &\multicolumn{3}{c}{\textbf{Traveled distance}~$\textbf{(m)}\uparrow$} 
        \\ 
        
        &
        &\multicolumn{1}{c}{\textbf{Rough}} 
        &\multicolumn{1}{c}{\textbf{Discrete}}
        &\multicolumn{1}{c}{\textbf{Stairs}} \\
        \hline\hline

        \multicolumn{1}{c}{\multirow{4}{*}{Pronk}}  
        & $0$& $15.13\pm 3.61$ & $17.34\pm 6.41$ & $13.56\pm 2.53$  \\
        & $0.25$ & $15.82\pm 3.43$  & $16.13\pm 6.65$  & \colorbox{lightgray}{$14.78\pm 2.69$}  \\
        & $0.5$ & $15.59\pm 3.38$  & \colorbox{lightgray}{$18.42\pm 6.29$}  & $14.06\pm 2.74$  \\
        & $0.75$ & \colorbox{lightgray}{$16.98\pm 4.05$}  & $16.79\pm 6.64$  & $14.07\pm 5.13$ \\
        \hline

        \multicolumn{1}{c}{\multirow{4}{*}{Trot}}  
        & $0$& $16.72\pm 3.53$ & $16.62\pm 6.50$ & $14.92\pm 2.82$  \\
        & $0.25$ & $14.98\pm 4.03$  & \colorbox{lightgray}{$18.96\pm 6.14$}  & $15.94\pm 2.86$   \\
        & $0.5$ & \colorbox{lightgray}{$16.84\pm 3.72$}  & $18.29\pm 6.18$  & \colorbox{lightgray}{$16.34\pm 2.41$}  \\
        & $0.75$ & $15.32\pm 3.04$  & $15.59\pm 5.89$  & $15.70\pm 2.32$  \\
        \hline

        \multicolumn{1}{c}{\multirow{4}{*}{Pace}}  
        & $0$& $17.11\pm 3.51$ & $17.48\pm 5.89$ & $16.83\pm 2.35$   \\
        & $0.25$ & \colorbox{lightgray}{$18.48\pm 4.45$}  & $18.43\pm 6.58$  & $16.71\pm 3.40$  \\
        & $0.5$ & $16.31\pm 3.95$  & \colorbox{lightgray}{$19.49\pm 5.77$}  & \colorbox{lightgray}{$18.18\pm 2.49$}\\
        & $0.75$ & $17.23\pm 4.19$  & $18.05\pm 5.67$  & $16.07\pm 4.15$ \\
        \hline

        \multicolumn{1}{c}{\multirow{4}{*}{Bound}}  
        & $0$& $14.57\pm 2.91$ & $14.53\pm 5.64$ & $14.17\pm 2.29$  \\
        & $0.25$ & \colorbox{lightgray}{$16.99\pm 3.98$}  & $16.74\pm 5.44$  & \colorbox{lightgray}{ $15.53\pm 2.55$}  \\
        & $0.5$ & $16.04\pm 3.57$  & $17.43\pm 5.16$  & $15.48\pm 2.92$ \\
        & $0.75$ & $16.66\pm 4.40$  & \colorbox{lightgray}{$17.47\pm 6.33$}  & $14.39\pm 4.13$  \\
        \hline

        \multicolumn{1}{c}{\multirow{4}{*}{\begin{tabular}[c]{@{}c@{}}Rotary\\gallop\end{tabular}}}  
        & $0$& $16.67\pm 3.56$ & $16.71\pm 6.20$ & $16.40\pm 2.88$  \\
        & $0.25$ & $17.89\pm 4.15$  & $17.77\pm 5.69$  & $17.17\pm 3.02$  \\
        & $0.5$ & \colorbox{lightgray}{$18.18\pm 3.91$}  & \colorbox{lightgray}{$18.61\pm 5.08$}  & \colorbox{lightgray}{$18.04\pm 2.10$}  \\
        & $0.75$ & $16.97\pm 4.68$  & $18.55\pm 5.88$  & $16.38\pm 3.94$  \\
        \hline
    
    \end{tabular}
    \begin{tablenotes}
        \footnotesize
        \item The average and standard deviation of the traveled distances are reported for rough, discrete, and stairs terrains in simulation using $1,\!000$ rollouts. The shaded values indicate the best performance for each terrain.
    \end{tablenotes}
    \end{threeparttable}
\end{table}

We measured the average traveled distance for different gaits with and without compliance threshold to assess the trade-off between accuracy and robustness in the proposed compliant contact tracking method. Table~\ref{table:gait_tracking} summarizes the results obtained from simulation using $1,\!000$ rollouts. For each robot, we set a command velocity of $1.2~\mathrm{m/s}$ and a gait cycle period of $0.4~\mathrm{s}$. The maximum episode length of the simulation was set to $20~\mathrm{s}$. 

The traveled distance of the robot consistently increases when using the compliant tracking method compared with the baseline~($\delta\!=\!0$)~\cite{margolis2023walk}. This performance gain is attributed to the compliance introduced in the foot contact tracking, which allows the robot to adapt to follow the desired gait pattern, but also comply to violate the gait pattern when necessary to overcome obstacles. The performance improvement is more pronounced on discrete and stairs terrains, where the robot is more vulnerable to external disturbances due to frequent foot collisions with the terrain.

\begin{figure}[!t]
	\centering 
	\begin{subfigure}[b]{0.95\linewidth}
		\includegraphics[width=\linewidth]{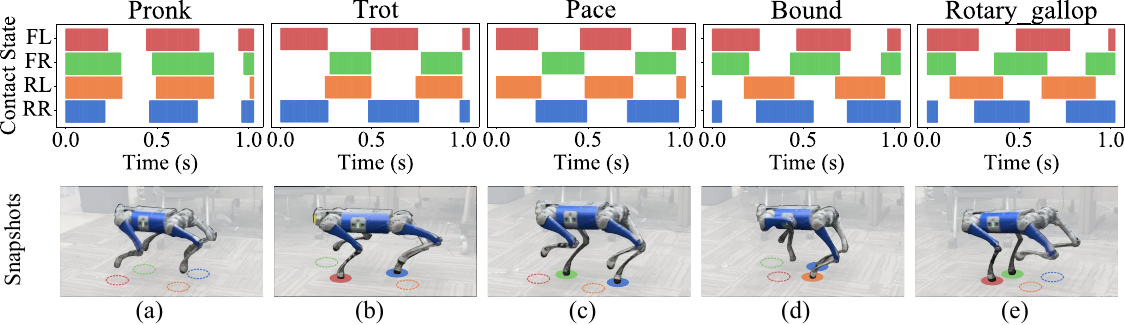}
	\end{subfigure}
	\caption{Gait tracking performance of the locomotion policy. The top row shows the foot contact states, while the bottom row displays snapshots of the robot. The robot accurately follows the desired gait patterns for various gaits, namely: (a)~pronk, (b)~trot, (c)~pace, (d)~bound, and (e)~rotary gallop.}
	\label{figure:gait_plot}
\end{figure}

The proposed compliant contact tracking allows our robot to track foot contact states accurately. A qualitative performance evaluation is presented in Fig.~\ref{figure:gait_plot}. The top row displays the foot contact states, while the bottom row shows snapshots of the robot for five different gaits. The robot accurately and robustly tracks the desired foot contact positions for all gaits.
\vspace{-2mm}


\subsection{Motion Description Data Scaling}\label{section:results_reasoning}
\begin{figure}[!t]
    \vspace{-10pt}
	\centering 
    \includegraphics[width=\linewidth]{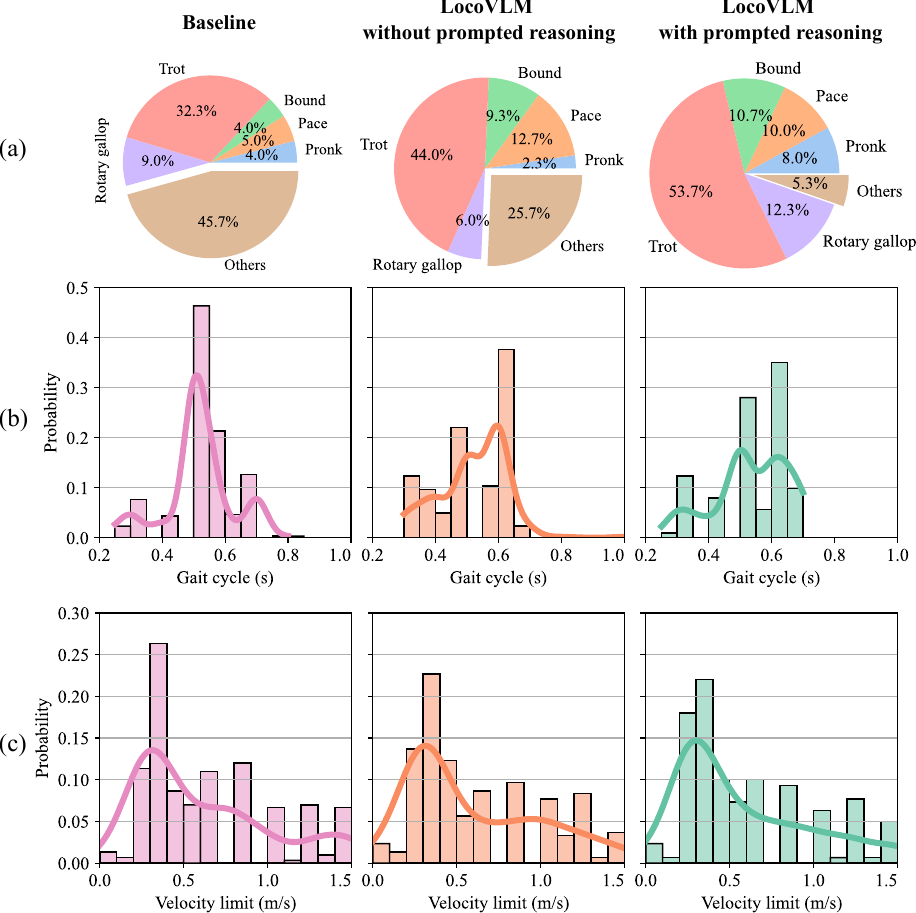}
	\caption{Statistics of the skill database generated with baseline~\cite{tang2023saytap}, \acro without prompted reasoning, and \acro with prompted reasoning. Each database contains $300$ data points and was generated using the same LLM-generated instructions. The statistics include (a)~categorical gait distribution, (b)~ histograms of gait cycle period distribution, and (c)~histograms of velocity limit distribution. The histograms are normalized to have a total probability of $1$.}
	\label{figure:statistics}
    \vspace{-10pt}
\end{figure}
We compare the skill database statistics of $300$ data points generated by the LLM in Fig.~\ref{figure:statistics}. As a baseline, we generated \motdescs using a method similar to SayTap~\cite{tang2023saytap}, where the LLM is prompted to generate one \motdesc at each step without any reasoning. We then compare this baseline with variants of \acro: one without prompted reasoning and another with prompted reasoning.

The key difference between the baseline and \acro without prompted reasoning is how the \motdescs are generated. \acro queries the VLM to generate \motdescs in batches, significantly reducing computational and monetary costs. For instance, the total cost of generating $300$ \motdescs using the baseline, \acro without prompted reasoning, and \acro with prompted reasoning is approximately $1.16$, $0.21$, and $0.25$ USD, respectively, when using the GPT-4o model. Furthermore, this batch generation process provides the LLM with a form of memory, preventing duplicate \motdescs from being generated across queries.

\subsubsection{Categorical Gait Distribution}
As shown in Fig.~\ref{figure:statistics}(a), the baseline database contains many unstructured gait phase offsets (marked as \textit{others}), which may compromise robot stability. \acro without prompted reasoning reduces these unstructured gaits by generating \motdescs in batches, decreasing repetition and instability. Prompted reasoning further improves the distribution, reducing unstructured gaits and promoting a more even spread across the five standard gaits. This structured understanding is especially beneficial for vague commands lacking explicit gait cues, such as ``\lang{shh! someone is sleeping, move quietly}''.

\subsubsection{Motion Descriptors Statistics}
The histogram in Fig.~\ref{figure:statistics}(b) shows that prompted reasoning leads to a more balanced distribution of gait cycle periods within $0.2$ to $0.7~\textrm{sec}$. In contrast, the baseline and \acro without reasoning cluster around $0.5~\textrm{sec}$ and include more outliers, with unstable values up to $T = 1.0~\textrm{sec}$. As for the velocity limit (Fig.~\ref{figure:statistics}(c)), there is no notable difference across methods, likely because this parameter is more intuitively inferred from instructions than the gait period or phase offset.

\subsection{Retrieval Performance}\label{section:results_retrieval}
\subsubsection{Accuracy} To quantitatively evaluate the retrieval performance of the VLM, we manually annotated $100$ instructions and their corresponding \motdescs based on our understanding of the instructions due to the lack of publicly-available groundtruth. 

\begin{table}[!t]
    \vspace{-12pt}
    \centering
    \footnotesize
    \caption{Retrieval accuracy of \acro using different retrieval metrics for $100$ instructions from the database.}
    \label{table:retrieval_accuracy}
    \resizebox{\linewidth}{!}{%
        \begin{tabular}{lcc|c}
        \hline
        
        \textbf{Retrieval Metric} & \textbf{Text as String} & \textbf{Text as Image} & \textbf{Average}\\
        \hline\hline

        Cosine similarity   & $21/100$ & $30/100$ & $20.5\%$\\
        Top-$K$ similarity  & $27/100$ & $48/100$& $37.5\%$\\
        Top-$K$ to ITM      & $51/100$ & $57/100$ & $54.0\%$\\
        Mixed-precision     & $\mathbf{72/100}$ & $\mathbf{87/100}$ & $\mathbf{79.5}\%$\\
        \hline
        
        \end{tabular}
    }
\end{table}
We hypothesize that retrieval performance degrades as the instruction database scales, which is supported by the results in Table~\ref{table:retrieval_accuracy}. With the text as string input, cosine similarity correctly retrieved only $21\%$ of the instructions. The top-$K$ similarity improves this to $27\%$ by narrowing the retrieval scope to the most similar candidates. Re-ranking with the BLIP-2 ITM head increases accuracy to $51\%$, but the improvement remains marginal.

Our proposed mixed-precision retrieval significantly improves performance, retrieving $72\%$ of the instructions, suggesting greater robustness and accuracy. The improvement is attributed to combining two levels of similarity: top-$K$ similarity captures low-level textual cues, while the ITM metric captures higher-level semantics. Their combination enables more accurate retrieval than either metric alone. Additionally, incorporating text-as-image representation further improves the accuracy up to $87\%$.

\subsubsection{Interpretation of Queries Out of the Database.}\label{section:results_interpretation}
One important feature of \acro is its ability to interpret instructions outside of the database. This enables users to interact with the robot more naturally and intuitively, without being constrained by the queries in the database. For instance, the VLM interprets the query ``\lang{you are a kangaroo}'' as ``\lang{let's jump like a rabbit}'', even though the query is absent from the database. Similarly, it responds to the query ``\lang{this is a library!}'' by suggesting ``\lang{move quietly}'', which aligns with common behavior in a library. These results highlight that the semantic reasoning capabilities of the VLM are sufficient to retrieve feasible motion commands, even for out-of-database queries.

\subsubsection{Interpretation of Robot-Centric Image Queries.}\label{section:results_robotimage}

\begin{figure}[!t]
	\centering 
	\includegraphics[width=\linewidth]{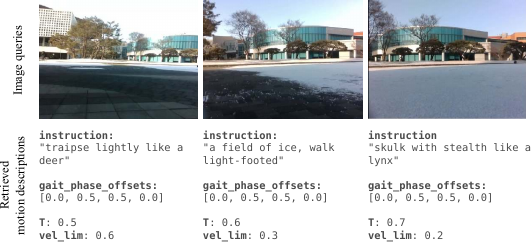}
	\caption{\acro succesfully interprets the scenes in the form of RGB images and provides the robot with \motdescs that are suitable for the observed scenes.}
	\label{figure:exp_outdoor}
\end{figure}

We evaluated \acro's performance on image-based queries captured by the robot's onboard camera without additional processing as. The experiment took place outdoors in a campus environment, where the robot transitioned from pavement to snow-covered terrain as shown in Fig.~\ref{figure:exp_outdoor}.

On the pavement, \acro interpreted the scene as ``\lang{traipse lightly like a deer}'', prompting a trot with a moderate velocity limit of $0.6~\text{m/s}$. In snowy areas, it produced interpretations like ``\lang{a field of ice, walk light-footed}'' or ``\lang{skulk with stealth like a lynx}'', resulting in slower gaits with lower velocity limits ($0.2$-$0.3~\text{m/s}$) and longer gait cycle periods ($0.6$-$0.7~\text{s}$). The lynx analogy is especially notable given the lynx's natural habitat in cold, snowy environments.

These results demonstrate \acro's ability to process robot-centric image queries and generate context-aware locomotion commands. This is particularly useful when environments are visually distinct but geometrically similar, as in our pavement-to-snow scenario.

\subsection{Zero-Shot Generalization Across Embodiments}\label{section:results_humanoid}
We demonstrate the feasibility of utilizing \acro to generalize across different legged robot embodiment, specifically on a humanoid robot. We trained a style-conditioned locomotion policy for a Unitree H1 robot with minimum modifications, i.e., by changing the gait phase offsets only for two legs. Therefore, the policy is trained only on the \texttt{trot} and \texttt{pronk} gaits. 

\begin{figure}[!t]
	\centering 
	\begin{subfigure}[b]{0.98\linewidth}
		\includegraphics[width=\linewidth]{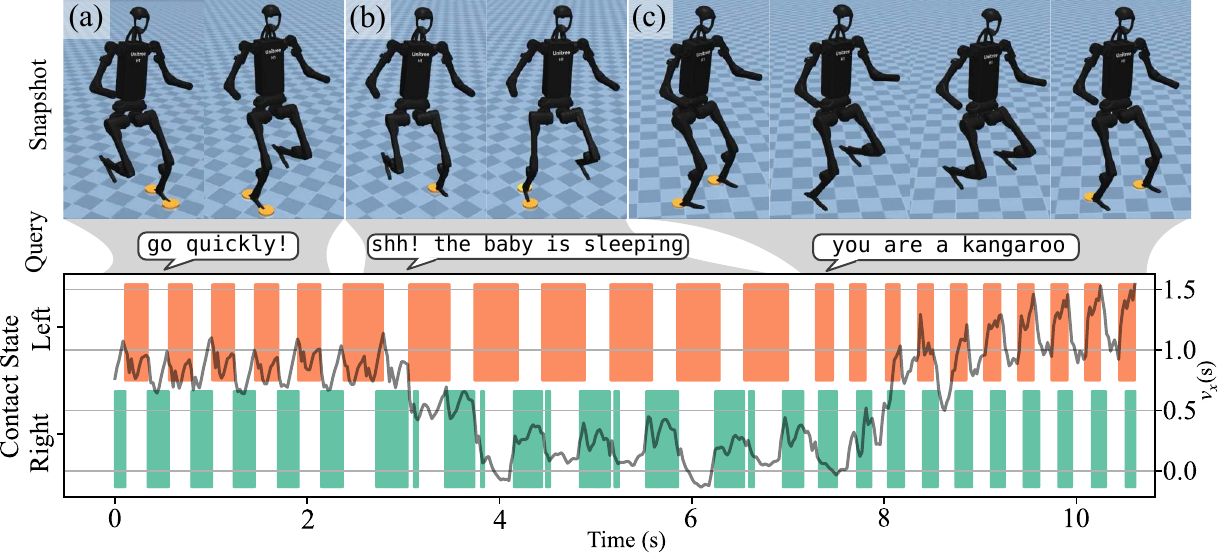}
	\end{subfigure}
	\caption{\acro can generalize to a humanoid robot in a zero-shot manner by utilizing only a subset of the gait phase offsets. The humanoid robot successfully adapts its locomotion style to follow the given text query.}
	\label{figure:humanoid}
\end{figure}

We utilized the skill database generated for quadrupedal robots by using only the first two gait offsets for each instruction and experimented in a MuJoCo simulation~\cite{todorov2012mujoco, menagerie2022github}. The results in Fig.~\ref{figure:humanoid} show successful skill retrieval and execution on the humanoid robot. This experiment demonstrates the generalizability of \acro that is attributed to the generalized motion parameterization and the instruction-grounded skill database.

\section{Conclusion}~\label{section:conclusion}
    In this paper, we presented \acro, a novel hierarchical locomotion adaptation framework that leverages pre-trained foundation models as a high-level motion advisor. Unlike recent works that utilize LLMs for in-the-loop control, \acro does not require in-the-loop LLM access, thereby enabling deployment in real-time systems without the need for continuous connectivity to the LLM. The proposed two-stage data generation process efficiently scales up data generation and distills the knowledge from the LLM into an efficient offline VLM. Our experiments demonstrate LocoVLM's skill expressiveness and generalizability across different legged robot embodiments and tasks. However, the current \acro interprets image and text separately, limiting the robot's ability to understand its surroundings through images while adhering to verbal instructions from human operators. Future work could focus on integrating text and image modalities to enhance scene understanding and task execution.

\bibliographystyle{IEEEtran}
\bibliography{main,utils/IEEEabrv}

\makeatletter
\newcommand{\forceappendixanchors}{%
    \setcounter{section}{0}
    \renewcommand{\theHsection}{appendix.\Alph{section}}%
}
\makeatother

\phantomsection
\forceappendixanchors

\newcommand{\bigappendix}{%
    \section*{\centering \Large Appendix}%
    \addcontentsline{toc}{section}{Appendix}%
}

\bigappendix

\section{Instruction Prompt}\label{appendix:instruction_prompt}
The instruction prompt is used to categorically generate three types of instructions: (1) mimicking behaviors, (2) scene descriptions, and (3) direct instructions. A template prompt is used to give general instructions to the LLM about its basic tasks, as shown in Listing~\ref{lst:system_prompt}. Afterwards, the instructions for each category are generated by substituting categorical prompts in Listings~\ref{lst:mimic_prompt}--\ref{lst:scene_prompt} into the template prompt.


\begin{lstlisting}[
    basicstyle=\ttfamily\scriptsize, 
    breaklines=true,
    caption={\small Template prompt for generating short instructions.},
    label={lst:system_prompt}
]
<General instructions>
#You are an expert in biomechanics, especially in quadrupedal biomechanics.
#You are familiar with various gait pattern of a quadrupedal animal.
#Given an input <N> that indicates the number for descriptions, you should output <N> different descriptions
{DESCRIPTION_TYPE}
\end{lstlisting}

\begin{lstlisting}[
    basicstyle=\ttfamily\scriptsize, 
    breaklines=true,
    caption={\small Categorical prompt for generating instructions to mimic certain behaviors.},
    label={lst:mimic_prompt}
]
Your task now is to tell a quadrupedal animal about how to:
- mimic an animal
- sometimes define the appropriate speed: slow, fast, moderate
- your output should be brief and concise for a short description

<Input-Output format and examples>
Input: 8
Output:
- let's hop like a rabbit
- run beautifully as a horse
- trot slowly, resembling a dog
- frogs jump by lifting all its foot!
- show me a hamster
- show me how a dog run quickly
- mimic a cat walk slowly
- do a galloping zebra!
\end{lstlisting}
\vspace{-5pt}
\begin{lstlisting}[
    basicstyle=\ttfamily\scriptsize, 
    breaklines=true,
    caption={\small Categorical prompt for generating instructions for giving directive commands.},
    label={lst:action_prompt}
]
Your task now is to tell a quadrupedal animal about how to:
- do some particular actions, especially related to:
* gait pattern
* foot stepping frequency
* speed
* loudness of the food stepping
* carefulness of the foot stepping

<Input-Output format and examples>
Input: 11
Output:
- trot with high frequency
- pace slowly and quietly
- trot slowly
- bound quickly!
- bound very slowly please
- tell me what a pace looks like
- oh no! catch that thief running!
- stop where you are!
- dance for me!
- reduce your noise
- let's be loud!
\end{lstlisting}
\vspace{-5pt}
\begin{lstlisting}[
    basicstyle=\ttfamily\scriptsize, 
    breaklines=true,
    caption={\small Categorical prompt for generating descriptions of particular scenes},
    label={lst:scene_prompt}
]
Your task now is to:
- Decribe a scene that can be encountered in daily lives
- The scene requires the animal gait to be changed, e.g. 
1. slowing down in a library
2. careful on ice
3. stopping at dangerous edge
- The description should be diverse but conceise

<Input-Output format and examples>
Input: 8
Output:
- be careful of the slippery tiles
- this is a library
- someone is sleeping. be quiet
- ouch! the floor is too hot!
- carefully approach the stairs
- on no! its a dead end!
- there is a meeting, silent
- the snow is slippery
\end{lstlisting}

\section{Skill Prompt}~\label{appendix:skill_prompt}
The skill prompt is used to translate instructions into \motdescs. The meta-prompt for the skill prompt is shown in Listing~\ref{lst:skill_prompt}. This prompt tells the LLM to generate motion descriptors based on the input instructions. Before generating the motion descriptors, the LLM is first required to generate a reasoning prompt explaining the rationale behind the motion descriptors~(Section~\ref{section:method_reasoning}). The motion descriptors are generated in the form of a structured \texttt{.json} file content. 

We also shuffled the input instructions to the LLM to prevent the model from memorizing the input-output pairs. This shuffling process is crucial to prevent the LLM from generating repetitive motion descriptors.

\begin{lstlisting}[basicstyle=\fontfamily{\ttdefault}\scriptsize, 
breaklines=true,
caption={Meta-prompt for generating motion descriptors given a list of instructions.}\label{lst:skill_prompt}
]
<General description>
You are a reinforcement learning engineer expert, specializing in quadrupedal robots.
Your task it to provide gait style commands of a quadrupedal robot based on the user input that describes the robot's behavior or surroundings.
You are required to ALWAYS give the output in a correct and standardized form.

<What is gait styles?>
We  parameterize gait styles using 2 parameterized values
1. T: float  
 - T is the gait cycle duration of the robot
 - T is proportional to 1/f, where f is the number of foot contact per 1 s 
 - 0.2 < T <= 1
 - Lower T means more step per second and vice versa
 - Higher T is required for longer walking stride similar to how cheetah runs.

2. gait_phase_offsets: List[float, float, float, float] 
 - The sequence of the variable in the list is [FL, FR, RL, RR]
 - FL=front left foot
 - FR=froont right foot
 - RL=rear left foot
 - RR=rear right foot
 - each offsets ranges between 0 to 1

3. vel_lim: float
 - The robot can ideally move up to 1.5m/s
 - The robot is small, its body length is around 40cm
 - Setting low vel_lim will make the robot move slower
 - Setting high vel_lim allows the robot to move faster
 - Set vel_lim to 0 to make the robot completely stop


<Some basic gait_phase_offsets dict>
You can use this dict as your reference. But I expect you to be creative, not confined to this example only.
gait_dict = {
            'trot': [0.0, 0.5, 0.5, 0.0],
            'rotary_gallop': [0.0, 0.2, 0.7, 0.5],
            'pace': [0.0, 0.5, 0.0, 0.5],
            'pronk': [0.0, 0.0, 0.0, 0.0],
            'bound': [0.0, 0.0, 0.5, 0.5],
        }
A brief contextual description:
1. Trot is a diagonal foot contact, known to be one of the stable gait, but not so creative
2. Rotary gallop shows rotary contact for each foot. 
3. Pace shows synchronous contacts between each left/right hemisphere
4. Pronk lifts all feet at the same time and land at the same time
5. Bound lifts all front or all rear legs at the same time
6. Gallop, pronk, and bound are typically more efficient with higher T (>0.5) on higher speed because of more leg flying phase.

<Important things to note>
1. Your output MUST NOT biased or overfit to those in the gait dict. The gait dict is only an example gaits, where the controller can perform well. You are free to create new gait lists and be creative!
2. You can select T between 0 to 1, but in general, 0.3<=T<=0.6 is preferrable for stability. There could be cases where you need T>0.6, but it is not desired. 
3. NEVER give T<0.2

<Reasoning>
You should output a brief reasoning of the gait that you recommend.

<Example input-output format>
Input: 
-trot slowly
-trot quickly
-pace quickly
-save your energy
-bound like a rabbit
-walk slowly
-you are a frog
-gallop swiftly
-be quiet!
Output: 
{
    "instruction": "trot slowly",
    "reason": "trot with low vel_lim",
    "T": 0.6,
    "gait_phase_offsets": [0.0, 0.5, 0.5, 0.0],
    "vel_lim": 0.4
}, 
{
    "instruction": "trot quickly",
    "reason": "trot with high vel_lim",
    "T": 0.3,
    "gait_phase_offsets": [0.0, 0.5, 0.5, 0.0],
    "vel_lim": 1.2
},
{
    "instruction": "pace quickly",
    "reason": "pace gait, high vel_lim, low T for high frequency",
    "T": 0.35,
    "gait_phase_offsets": [0.0, 0.5, 0.0, 0.5],
    "vel_lim": 1.0
},
{
    "instruction": "save your energy",
    "reason": "pace gait energy-efficient, reduce to moderate vel_lim and T",
    "T": 0.4,
    "gait_phase_offsets": [0.0, 0.5, 0.0, 0.5],
    "vel_lim": 0.5
},
{
    "instruction": "bound like a rabbit",
    "reason": "bound with moderate vel_lim, increase T for longer swing",
    "T": 0.6,
    "gait_phase_offsets": [0.0, 0.0, 0.5, 0.5],
    "vel_lim: 1.0
    
},
{
    "instruction": "walk slowly",
    "reason": "make a slow gallop gait, reduce vel_limit and increase T for lower step noise",
    "T": 0.6,
    "gait_phase_offsets": [0.0, 0.2, 0.7, 0.5],
    "vel_lim: 0.3
    
},
{
    "instruction": "you are a frog",
    "reason": "frog jump with all legs flying, set moderate vel_lim, set moderate T for moderate swing time",
    "T": 0.45,
    "gait_phase_offsets": [0.0, 0.0, 0.0, 0.0],
    "vel_lim": 0.6
},
{
    "instruction": "gallop swiftly",
    "reason": "make a gallop gait, increase T for flying phase, increase vel_lim for running",
    "T": 0.6,
    "gait_phase_offsets": [0.0, 0.2, 0.7, 0.5],
    "vel_lim: 1.2
    
},
{
    "instruction": "be quiet!",
    "reason": "gallop slowly with high T for lower noise",
    "T": 0.65,
    "gait_phase_offsets": [0.0, 0.2, 0.7, 0.5],
    "vel_lim: 0.25   
}
\end{lstlisting}
\section{Versatile Locomotion Implementation Details}\label{appendix:versatile_implementation}
\subsection{Foot Phase Offsets}\label{appendix:foot_phase}
The policy was trained using five representative gaits: pronk, trot, pace, bound, and rotary gallop. While other gaits can also be learned, these five were selected as they represent fundamental locomotion styles that can be parameterized by the foot phase offsets. The foot phase offsets for each gait are summarized in Table~\ref{table:gait_offsets}, with each set defined for the front left, front right, rear left, and rear right feet, respectively.

\begin{table}[h!]
    \centering
    \vspace{-4mm}
    \caption{Foot phase offsets for each gait style. FL, FR, RL, and RR represent the front left, front right, rear left, and rear right feet, respectively.}
    \footnotesize
    \begin{tabular}{llcccc}
        \hline
        \multicolumn{1}{c}{\multirow{2}{*}{\textbf{Gait}}} & \multicolumn{4}{c}{\textbf{Phase offsets}} \\
        &FL&FR&RL&RR\\
        \hline\hline
        Pronk & 0.0 & 0.0 & 0.0 & 0.0 \\
        Trot & 0.0 & 0.5 & 0.5 & 0.0 \\
        Pace &  0.0 & 0.5 & 0.0 & 0.5 \\
        Bound & 0.0 & 0.0 & 0.5 & 0.5 \\
        Rotary gallop & 0.0 & 0.2 & 0.7 & 0.5 \\
        \hline
    \end{tabular}
    \label{table:gait_offsets}
\end{table}

\subsection{Compliant Contact Tracking Parameters}\label{appendix:contact_tracking}
The compliance threshold in Eq.~\ref{equation:compliance} was set to $\delta=0.5$ for all experiments, allowing inaccurate contact states within $50\%$ of the gait cycle. While the compliance threshold may seem large at first glance, it is crucial to ensure that the locomotion controller can recover from disturbances while maintaining the desired gait style.

The \gaitencoding vector is concatenated with the observation vector before being fed into the policy network. The compliance zone for foot contact tracking is defined as $-0.5 \leq \phi(t) \leq 0.5$ to provide greater flexibility. 

The contact tracking error reward is smoothed using an exponential kernel function, similar to the approach applied to the linear velocity tracking term in~\cite{miki2022learning,rudin2022learning}. The exponential kernel function for the contact tracking error is defined as:
\begin{equation}
    r_\text{contact} = \exp\left(-\frac{\phi_\text{error}}{\sigma}\right),
\end{equation}
where $\sigma$ is the smoothing factor, set to $0.25$ for all experiments. The compliant contact tracking method allows the locomotion controller to recover from disturbances while preserving gait styles whenever possible.

\begin{figure}[!ht]
	\centering 
	\includegraphics[width=\linewidth]{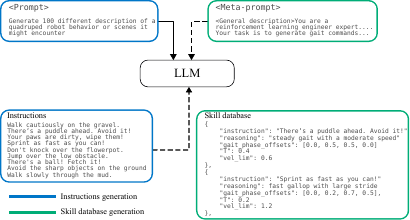}
	\caption{Offline skill database generation pipeline. The LLM firstly generates instructions, which are categorized into mimicking behaviors, scene responses, and direct instructions. These instructions are then passed to a meta-prompt to generate contents for the skill database.}
	\label{figure:datagen}
	\vspace{-5pt}
\end{figure}
The two-stage description generation as shown in Fig.~\ref{figure:datagen} offers two primary benefits: (1) it mitigates the maximum token length constraint of the LLM, and (2) it enables the LLM to generate diverse and structured motion descriptors by passing the generated instructions to a meta-prompt that produces the motion descriptors. We used the \llm model as the LLM.

Rather than generating all three types of instructions in a single prompt, we observed that the LLM produces more diverse yet structured instructions when prompted separately for each category. A crucial key implementation detail was prompting the LLM to explicitly generate $n$ instructions for each category instead of generating all instructions simultaneously. This categorical prompting approach prevents the LLM from hallucinating false instructions and ensures that the generated instructions are relevant to the category. To comply with the maximum token length constraint of the LLM, we set $n\!=\!100$ for each generation process.

\section{Details of Mixed-Precision Retrieval}\label{appendix:mixed_retrieval}
In the first stage, we retrieve the top-$K$ instructions using the cosine similarity between the query and the instruction embeddings. This stage is formulated as follows:
\begin{equation}
	\mathbf{I}^K = \argmax_{\mathcal{I} \in \mathcal{D}} \cossim \big(f_\text{BLIP}(\mathcal{I}_\text{query}), f_\text{BLIP}(\mathbf{I})\big),
\end{equation}
where $\cossim(.,.)$ is the cosine similarity operator, $f_\text{BLIP}$ denotes the BLIP-2 model encoder, and $\mathbf{I}^K$ represents the top-$K$ retrieved instructions. We store the probability of a positive match between the query and the top-$k$ instructions as:
\begin{equation}
	p_\text{1}(\mathbf{I}^K) = \softmax \left(\cossim \big(f_\text{BLIP}(\mathcal{I}_\text{query}), f_\text{BLIP}(\mathbf{I}^K)\big)\right).
\end{equation}
The probability computation is performed only on $\mathbf{I}^K$ and not the entire database. This choice is based on the observation that computing probabilities over the whole database yields similar values for each instruction, making the probabilities less informative.

In the second stage, we use the ITM head of the BLIP-2 model to re-rank the top-$K$ instructions. The ITM head computes the positive and negative match probabilities between $\mathcal{I}_\text{query}$ and each instruction $\mathcal{I}_k \in \mathbf{I}^K$. The probability of a positive match between $\mathcal{I}_\text{query}$ and each $\mathcal{I}_k$ is computed as follows:
\begin{equation}
	p_\text{2}(\mathcal{I}_k) = \softmax \big( f_\text{ITM}(\mathcal{I}_\text{query}, \mathcal{I}_k) \big),
\end{equation}
where $k$ is the index of the top-$K$ instructions, $f_\text{ITM}(.,.)$ denotes the BLIP-2 model's ITM inference.

Finally, the retrieved instruction is selected as the one with the highest probability from the mixed-precision retrieval process. Substituting the total probability into Eq.~\ref{eq:retrieval}, we obtain:
\begin{equation}
	\mathcal{I}^* = \argmax_{\mathcal{I}_k} \left( p_\text{1}(\mathbf{I}^K) + p_\text{2}(\mathbf{I}^K) \right).
\end{equation}

\section{Output of Reasoning Prompts in Motion Descriptors Generation}~\label{appendix:results_reasoning}
We qualitatively evaluate several examples of generated reasonings, as shown in Table~\ref{table:reasoning}. In general, the generated reasonings translate the instructions into more detailed and technical descriptions, which are particularly useful when the instructions are vague. Overall, we observe that prompted reasoning enables the LLM to generate a more diverse range of skills while maintaining the accuracy and feasibility of the generated \motdescs.
\begin{table}[!t]
    \centering
    \footnotesize
    \caption{Instructions and their corresponding reasonings generated by the LLM.}
    \label{table:reasoning}
    \begin{center}
    \begin{tabular}{m{0.4\linewidth}m{0.48\linewidth}}
    
    \hline
    \multicolumn{1}{c}{\textbf{Instructions}} & \multicolumn{1}{c}{\textbf{Reasonings}} \\
    \hline\hline

    \lang{trundle along like a hippo} & \lang{slow and heavy trot, lower vel\_lim and increase T} \\ \hline
    \lang{oh no! catch that thief running!} & \lang{fast and aggressive gait, low T and high vel\_lim} \\ \hline
    \lang{the sound of a human voice, stay hidden.} & \lang{use a trot with high T for stealthy movement, low vel lim for quietness} \\ \hline
    \lang{a busy marketplace, navigate through the crowd.} & \lang{slow pace with moderate T for careful navigation} \\ \hline
    
    \end{tabular}
    \end{center}
\end{table}

\section{Robot-Centric Images as Navigation Constraints}\label{appendix:results_navigation}

\begin{figure}[!t]
	\centering
	\begin{subfigure}[b]{0.48\textwidth}
		\includegraphics[width=1.0\linewidth]{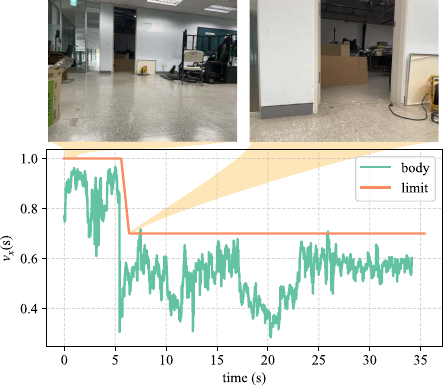}
	\end{subfigure}
	\caption{Changes in the velocity limit from \acro yield safer navigation in cluttered environments. The local planner adjusts the command velocity while respecting the limit set by \acro.}
	\label{figure:exp_clutter}
\end{figure}

\begin{figure}[!t]
	\centering
	\begin{subfigure}[b]{0.48\textwidth}
		\includegraphics[width=1.0\linewidth]{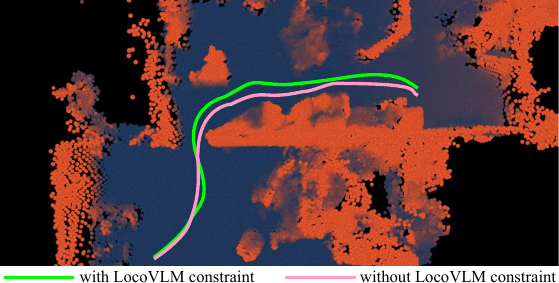}
	\end{subfigure}
	\caption{Comparison of navigation performance with and without \acro constraints. The 3D pointcloud map illustrates the density of obstacles in the environment, where obstacles and free space are colored with red and navy blue, respectively. The planner running with \acro constraints maintains a bigger distance-to-obstacle clearance, reducing the likelihood of collisions.}
	\label{figure:exp_clutter_compare}
\end{figure}

Beyond locomotion, we also explore how the proposed \acro framework can provide constraints to a navigation stack by utilizing a subset of the skill database. Specifically, we used the same skill database and employed \acro to retrieve only the velocity limit parameter. This parameter was then used to constrain the maximum allowable velocity in the robot's local planner. We evaluated the performance of the navigation stack with and without the \acro constraints in an environment where the robot needs transitions from wide to narrow and cluttered spaces. The VLM inference period was set to $5~\textrm{sec}$ to reduce jitter due to frequent velocity limit changes.

In this experiment, the robot starts in a wide indoor space and navigates to a goal point located in another room containing cluttered objects, resulting in a narrow space transition. The velocity limit recommendation from the \acro module, shown in Fig.~\ref{figure:exp_clutter}, demonstrates how the velocity limit changes as the robot enters the narrow space. This adjustment ensures the robot navigates with higher safety and sufficient time to react to obstacles in the cluttered room. Inside the cluttered room, \acro consistently advises the robot to navigate with a lower velocity limit to avoid collisions with obstacles.

We further compare the navigation performance with and without \acro constraints in Fig.~\ref{figure:exp_clutter_compare}. Using the same local planner, the system with constraints provided by \acro maintains a bigger distance-to-obstacle clearance. This result underscores the applicability of semantic-driven constraints in enhancing navigation stacks that rely solely on geometric information.

\end{document}